\documentclass[]{spie}  

\usepackage{amsmath,amsfonts,amssymb}
\usepackage{graphicx}
\DeclareMathOperator*{\argmin}{argmin}
\usepackage[colorlinks=true, allcolors=blue]{hyperref}

\title{Learning unbiased group-wise registration (LUGR) and joint segmentation: evaluation on longitudinal diffusion MRI}

\author[a]{Bo Li}
\author[a,b]{Wiro J. Niessen}
\author[a]{Stefan Klein}
\author[a,c,d]{M. Arfan Ikram}
\author[a,c]{Meike W. Vernooij}
\author[a]{\mbox{Esther E. Bron}}
\affil[a]{Department of Radiology and Nuclear Medicine, Erasmus MC, Rotterdam, the Netherlands}
\affil[b]{Imaging Physics, Applied Sciences, Delft University of Technology, Delft, the Netherlands}
\affil[c]{Department of Epidemiology, Erasmus MC, Rotterdam, the Netherlands}
\affil[d]{Department of Neurology, Erasmus MC, Rotterdam, the Netherlands}

\authorinfo{Further author information: (Send correspondence to B. Li)\\B. Li: b.li@erasmusmc.nl }

\begin{document} 
\maketitle

\begin{abstract}
Analysis of longitudinal changes in imaging studies often involves both segmentation of structures of interest and registration of multiple timeframes. The accuracy of such analysis could benefit from a tailored framework that jointly optimizes both tasks to fully exploit the information available in the longitudinal data. Most learning-based registration algorithms, including joint optimization approaches, currently suffer from bias due to selection of a fixed reference frame and only support pairwise transformations. We here propose an analytical framework based on an unbiased learning strategy for group-wise registration that simultaneously registers images to the mean space of a group to obtain consistent segmentations. We evaluate the proposed method on longitudinal analysis of a white matter tract in a brain MRI dataset with 2-3 time-points for 3249 individuals, i.e., 8045 images in total. The reproducibility of the method is evaluated on test-retest data from 97 individuals. The results confirm that the implicit reference image is an average of the input image. In addition, the proposed framework leads to consistent segmentations and significantly lower processing bias than that of a pair-wise fixed-reference approach. This processing bias is even smaller than those obtained when translating segmentations by only one voxel, which can be attributed to subtle numerical instabilities and interpolation. Therefore, we postulate that the proposed mean-space learning strategy could be widely applied to learning-based registration tasks. In addition, this group-wise framework introduces a novel way for learning-based longitudinal studies by direct construction of an unbiased within-subject template and allowing reliable and efficient analysis of spatio-temporal imaging biomarkers. 
\end{abstract}

\section{INTRODUCTION}
\label{sec:intro}  

The increasing availability of longitudinal imaging data is expanding our ability to characterize progressive anatomical changes, ranging from normal changes in the life span, to response along disease trajectories or therapeutic actions. Accurate analysis of longitudinal imaging changes could certainly benefit from a tailored analytical framework for data processing. Specifically, such frameworks are designed to combine structure segmentation and nonlinear registration to obtain within-subject anatomical consistency, to avoid processing bias during registration, and to make use of conditional correlations between segmentation and registration in longitudinal data by jointly optimizing the parameters. To date most existing frameworks have focused on parts of the above mentioned challenges, for instance, \cite{estienne2019u,li2019hybrid,qin2018joint} showed improved performances using joint optimization but the registration may be biased as fixed reference-frames were used. We here extend these approaches by enabling unbiased alignment among a given group of images, and thereby providing an analytical framework that can be used in longitudinal imaging studies for reliable quantification of biomarkers and efficient modeling of a within-subject template.

A common strategy in image registration is to choose a \textit{fixed} reference, to which the rest of images can be pair-wise registered. This approach however introduces a bias towards the selected reference, e.g., interpolation asymmetry \cite{reuter2012within}. In longitudinal studies, the baseline time-point usually serves as the \textit{fixed} reference, therefore only the follow-up images are smoothed and the baseline image is consistently used for initialization. While a possible solution is to consider all combinations of paired deformations, for instance, to compose them into a mean deformation \cite{seghers2004construction,bron2014diagnostic,huizinga2016pca} or to search for a median image \cite{park2005least,reuter2012within,vernooij2008white}, this results in a dramatic growth of computational complexity. An efficient strategy to eliminate the need and the bias of reference selection is to use an implicit reference image that has the \textit{mean} anatomy of the group. In classical registration algorithms, the \textit{mean} strategy has been widely applied \cite{balci2006free,bhatia2004consistent,metz2011nonrigid}. For learning-based registration, however, the \textit{fixed} strategy is still the current paradigm. A recently-proposed relevant idea is to learn an implicit \textit{mean-space} atlas using a loss term that penalizes the moving average of paired deformations to enforce geometric centrality \cite{dalca2019learning}. The use of a loss term however requires to explicitly set the weight for the penalty and thus to specify to what extent the constraint is satisfied \cite{bhatia2004consistent}. 

In this work, we propose a novel strategy for Learning-based Unbiased Group-wise Registration (LUGR) that (1) forces the implicit reference in the mean-space of the input images using a constraint on the velocity field, and (2) estimates simultaneously all invertible deformations between the inputs and the implicit reference. Also, by joint optimization with a segmentation network, the proposed framework (3) leads to perfectly consistent segmentation across images. We validate the method in longitudinal diffusion MRI analysis of white matter tracts, which describe functionally grouped axonal fibers. 

\section{New and breakthrough works to be presented}
\label{sec:new}
The main novelty and contributions of the presented work are as follows:
\begin{itemize}
  \item The idea of learning unbiased group-wise registration is novel and elegant compared to the existing pairwise strategies. To the best of our knowledge, this work is the first to use constraint optimization for learning a mean space in group-wise registration. This has impact beyond the presented application, as mean-space learning strategy can be widely applied to learning-based registration tasks.
  \item This framework introduces a novel way for learning-based longitudinal studies. Pairwise registration usually uses the baseline visit as fixed reference, suffering from an interpolation asymmetry bias \cite{reuter2012within}. Instead, our method directly constructs an unbiased within-subject template, allowing reliable and efficient analysis of imaging biomarkers.
\end{itemize}
\section{Methods}
\label{s:2}

\subsection{LUGR: Learning-based unbiased group-wise registration}
\label{ss:2.1}

Given a set of $n$ images $\mathcal{I} = \{I_1,\, ..., I_n \}$, each described by intensity values $I_i(x)$ at spatial coordinate $x \in X_i , i=1...n$, the group-wise registration aims to simultaneously estimate a set of transformations $\tau = \{\boldsymbol{\mathcal{T}}_1, ..., \boldsymbol{\mathcal{T}}_n\}$ that warp the images $\mathcal{I}$ to a common reference space $\overline X$, such that $I_i(\boldsymbol{\mathcal{T}}_i) \approx I_j(\boldsymbol{\mathcal{T}}_j) \, \forall i\neq j$.

To achieve invertiblity, we adopt diffeomorphic transformations \cite{ashburner2007fast} for $\tau$. Specifically, we estimate a set of stationary velocity fields (SVF): $\mathcal{V}=\{\boldsymbol{v}_i\}_{i=1}^n$, pointing from $\overline X$ to $X_i$. The transformations $\tau$ and their inverse transformations $\tau^{-1} = \{\boldsymbol{\mathcal{T}}^{-1}_i\}_{i=1}^n$ can be estimated subsequently with the integration of the velocity field $\mathcal{V}$ and the negative velocity filed $-\mathcal{V}$ over unit time  \cite{ashburner2007fast,dalca2019unsupervised,krebs2018unsupervised}. In this setting, the transformation aligning any two images within the group ($\boldsymbol{\mathcal{T}}_{i \rightarrow j}$) is also available by the composition of $\boldsymbol{\mathcal{T}}_j$ and $\boldsymbol{\mathcal{T}}^{-1}_i$, i.e., $X_j = \boldsymbol{\mathcal{T}}_j \big(\boldsymbol{\mathcal{T}}^{-1}_i (X_i)\big)$.

To estimate the transformation between any image pair, learning-based registration algorithms use a shared high-dimensional mapping function $\mathcal{G}_{\boldsymbol{\Psi}}$, that is, $\boldsymbol{\mathcal{T}}_{i \rightarrow j} = \mathcal{G}_{\boldsymbol{\Psi}}(I_i, I_j)$. We here extend the formulation of $\mathcal{G}_{\boldsymbol{\Psi}}$ from pair-wise estimation to simultaneous estimation of a set of transformations $\tau$, i.e.,
\begin{equation}\label{eq1}
\tau \leftarrow \mathcal{V} \leftarrow \mathcal{G}_{\boldsymbol{\Psi}}(\mathcal{I}).
\end{equation}

For the purpose of unbiased registration, the method takes into account three constraints during the optimization. First, for monomodal imaging data, intensities at corresponding spatial locations are expected to be similar among images. We therefore use an intensity-based dissimilarity metric, i.e., mean squared error, as part of the registration component of the loss function ($\mathcal{L}_{reg}$) following the approach of \cite{metz2011nonrigid}:
\begin{equation}\label{eq2}
\mathcal{L}_{reg}(\mathcal{I},\tau) = \frac{1}{n} \sum_{i=1}^{n} \big( \overline I - I_i ( \boldsymbol{\mathcal{T}_i}) \big)^2,
\end{equation}
where $\overline I$ can be considered as an implicit reference image: $\overline I = 1/n \sum_{i=1}^{n} I_i( \boldsymbol{\mathcal{T}}_i )$.

Second, to force the reference image $\overline I$ in the geometric center of the group, we propose a constrained optimization inspired by classical registration algorithms \cite{balci2006free,bhatia2004consistent,metz2011nonrigid}. For this, we substract the average velocity field from each of the estimated velocity fields, resulting in the sum of the velocity fields to be zero $\forall x \in \overline X$. In the reminder of the paper this procedure is referred to as a constraint projection layer, which is formulated as 
\begin{equation}\label{eq3}
\boldsymbol{\hat{v}}_i  = \boldsymbol{v}_i - \frac{1}{n} \sum_{i=1}^{n} \boldsymbol{v}_i \quad \implies \quad \sum_{i=1}^{n} \boldsymbol{\hat v}_i = 0. 
\end{equation}

Third, to encourage smoothness of the estimated velocity fields, a diffusion regularization term $\mathcal{L}_{def}$ is included in the loss function:
\begin{equation}\label{eq4}
\mathcal{L}_{def}(\mathcal{V}) = \frac{1}{n} \sum_{i=1}^{n} \|\nabla\boldsymbol{v}_i \|\, _2^2.
\end{equation}
To summarize, the loss function of LUGR is composed of an intensity dissimilarity term (Eqs.~(\ref{eq2})) and a regularization term (Eqs.~(\ref{eq4})), while subject to the constraint of geometric centrality (Eqs.~(\ref{eq3})).

\subsection{Joint optimization framework}
\label{sss:2.2}

    \begin{figure} [ht]
    \begin{center}
    \includegraphics[width=\textwidth]{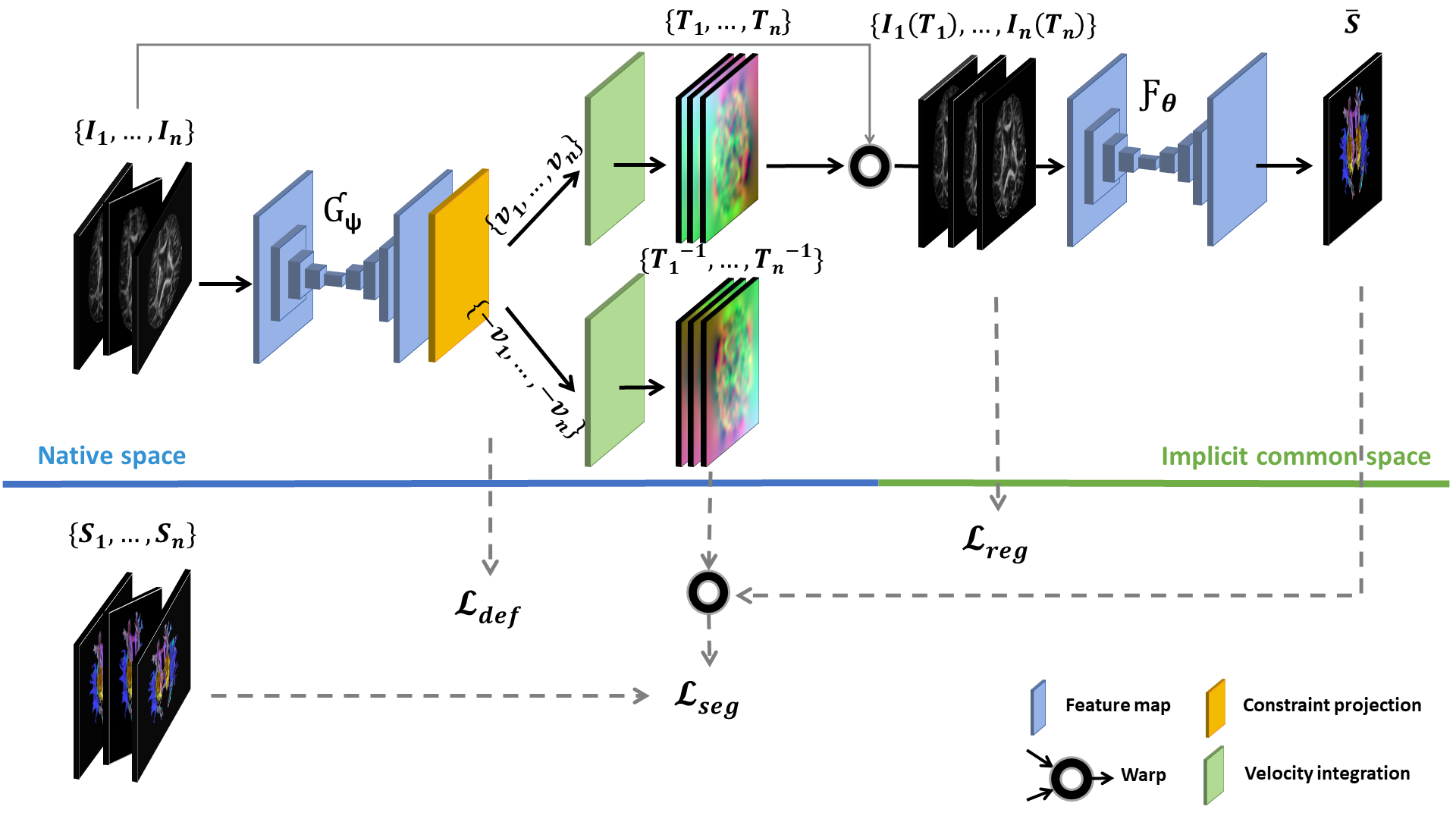}
    \setlength{\belowcaptionskip}{-15pt}
    \end{center}
\caption{\label{fig1}
Overview of the joint optimization framework. $\mathcal{G}_{\boldsymbol{\Psi}}$ and $\mathcal{F}_{\boldsymbol{\Theta}}$ denote the registration and segmentation network, respectively. The loss function consists of $\mathcal{L}_{seg}$, $\mathcal{L}_{reg}$ and $\mathcal{L}_{def}$ terms.}
    \end{figure}

In this work, we integrate the proposed unbiased registration module, LUGR, with a segmentation task in a single analytical framework. Input images are aligned to the mean-space to be jointly segmented, making use of information from all images and ensuring spatial correspondence of the segmented structure across images. The overview of the proposed joint framework is presented in Fig.~\ref{fig1}. 

In short, the image set $\{I_1,\, ..., I_n \}$ in native space are first aligned to an implicit mean-space of the group with $\mathcal{G}_{\boldsymbol{\Psi}}$. The aligned images $\{I_1(\boldsymbol{\mathcal{T}}_1),...,I_n(\boldsymbol{\mathcal{T}}_n)\}$ are then concatenated for the segmentation of $\overline S$, where $\overline S$ can be viewed as the segmentation of the implicit reference image $\overline I$. We assume that the structure of interest to be segmented has normal appearance. The segmentation task is implemented in a learning-based manner using a mapping function $\mathcal{F}_{\boldsymbol{\Theta}}$, i.e., 
\begin{equation}\label{eq5}
 \overline S = \mathcal{F}_{\boldsymbol{\Theta}}\big(I_1 ( \boldsymbol{\mathcal{T}}_1 ), ... , I_n ( \boldsymbol{\mathcal{T}}_n) \big).
\end{equation}
To optimize the parameters $\boldsymbol{\Theta}$ of $\mathcal{F}_{\boldsymbol{\Theta}}$, we warp the obtained segmentation $\overline S$ back to the image native space to minimize the classification error ($\mathcal{L}_{seg}$) between the warped segmentations $\{\overline S (\boldsymbol{\mathcal{T}}^{-1}_1),...,\overline S (\boldsymbol{\mathcal{T}}^{-1}_n) \}$ and the segmentation labels $\mathcal{S}=\{S_1,...,S_n\}$. This ensures spatial correspondence of segmentations across images and also optimizes the inverse transformations. The classification error is measured with a metric based on the weighted inner product ($w=3$) \cite{choi2010survey,li2020neuro4neuro}, which is formulated as:  
\begin{equation}\label{eq6}
\mathcal{L}_{seg}(\mathcal{S}, \overline S, \tau) = - \frac{1}{n} \sum_{i=1}^{n} \bigg(w S_i \times \overline S ( \boldsymbol{\mathcal{T}}^{-1}_i) + (1-S_i) \times \big( 1- \overline S ( \boldsymbol{\mathcal{T}}^{-1}_i) \big)\bigg)
\end{equation}

Combining Eqs.~(\ref{eq2}), (\ref{eq4}) and (\ref{eq6}), the loss function of the joint optimization framework $\mathcal{L}$ is defined as:
\begin{equation}\label{eq7}
  \mathcal{L} = \mathcal{L}_{reg}+ \lambda_1 \mathcal{L}_{def} + \lambda_2 \mathcal{L}_{seg},
\end{equation}
and optimized for $\boldsymbol{\Psi}$ and $\boldsymbol{\Theta}$ over all the $K$ sets of training samples $\{(\mathcal{I}^k, \mathcal{S}^k) \}_{k=1}^K$, i.e.,
\begin{equation} \label{eq8}
\hat{\boldsymbol{\Psi}},\hat{\boldsymbol{\Theta}} \leftarrow \argmin_{\boldsymbol{\Psi}, \boldsymbol{\Theta}} \sum_{k=1}^{K} \mathcal{L}_{reg}\big(\mathcal{I}^k, \mathcal{G}_{\boldsymbol{\Psi}}(\mathcal{I}^k)\big) + \lambda_1 \big( \mathcal{G}_{\boldsymbol{\Psi}}(\mathcal{I}^k)\big) + \lambda_2 \mathcal{L}_{seg} \big(\mathcal{S}^k, \mathcal{F}_{\boldsymbol{\Theta}}(\mathcal{I}^k), \mathcal{G}_{\boldsymbol{\Psi}}(\mathcal{I}^k) \big).
\end{equation}


We propose a generic framework where the architecture of the registration and segmentation components can be adapted based on specific applications. For the particular network architecture used in this study, we encoded $\mathcal{G}_{\boldsymbol{\Psi}}$ and $\mathcal{F}_{\boldsymbol{\Theta}}$ sequentially with modified U-Net \cite{ronneberger2015u} architectures. The convolution layers produce a set of $k$ feature maps by individually convolving the input with $k$ kernels. Here, we used $k=[16,32,64,128,256,128,64,32,16]$ for both $\mathcal{G}_{\boldsymbol{\Psi}}$ and $\mathcal{F}_{\boldsymbol{\Theta}}$. Each convolution layer was of kernel size $(3,3,3)$, followed by a leaky ReLu layer ($a = 0.2$) for non-linearity. Resamplings were performed using max-pooling and linear up-sampling layers. The last layers of $\mathcal{G}_{\boldsymbol{\Psi}}$ were convolution layers with 3 kernels that yielded stationary velocity fields $\boldsymbol{v}_i \in \mathbb{R}^{3\times3}$, then followed by the computation layers of constraint projection, velocity integration \cite{dalca2019unsupervised,krebs2018unsupervised}, and spatial transformation \cite{jaderberg2015spatial}. The last layer of $\mathcal{F}_{\boldsymbol{\Theta}}$ was an softmax function. For performance evaluation, the obtained probabilistic segmentation maps were binarized with a threshold of $P>0.5$.

\section{Experiments and results}
\label{s:3}
\subsection{Dataset and preprocessing}
\label{ss:3.1}

We applied the proposed method on 8045 diffusion-weighted MRI (DWI) scans of 3249 individuals from the Rotterdam Study \cite{hofman2015rotterdam}. Data from a set of 97 normal individuals who were scanned twice within a month was used for evaluations of reproducibility of the method, in which no changes in brain microstructure are expected. The scans from remaining individuals were repeatedly acquired in a time interval of 1–5 years, in which changes in brain microstructure are expected owing to aging. We combined these scans into a set of 6043 \textit{paired-scans} (two time-points, $n=2$) and an overlapping set of 1440 \textit{grouped-scans} (three time-points, $n=3$). We split the set of \textit{paired-scans} into $5175:200:668$ for training, validation and test; the split of the \textit{grouped-scans} is $1240:100:100$. The data split was based on the participants, i.e., we made sure that scans from the same participant ended up in either training, validation, test, or reproducibility dataset.

DWI was acquired at 1.5 Tesla with 25 diffusion-weighted volumes (b-value of $1000 s/mm^2$) and 3 non-weighted volumes. Each volume has $211\times210\times123$ voxels, of which the size is $1 mm^3$. DWI preprocessing \cite{li2019hybrid} included correction of motion and eddy currents, estimation of diffusion tensors, and computation of fractional anisotropy (FA) and mean diffusivity (MD). FA image, a scalar map measuring anisotropy degree of diffusion, was used as the input of the method. For each DWI in the dataset, reference segmentation of white matter tracts were available using the approach of \cite{de2015tract} that based on probabilistic tractography and atlas information. In the present work, we focused on the segmentation of forceps minor (FMI) tract and utilized the reference segmentations as labels for model training and performance evaluation.  

\subsection{Experiments and evaluation metrics}\label{ss:3.2}

\subsubsection{Reference methods}\label{sss:3.2.1} 
We compare the proposed mean-space joint framework ([Mean]) with a pair-wise framework \cite{li2019hybrid} that uses fixed-reference registration ([Fixed]). The [Fixed] method registers a moving image to a target image ($\mathcal{L}_{reg}^F$), encourages deformation smoothness ($\mathcal{L}_{def}^F$), and segments the moving image ($\mathcal{L}_{seg}^F$). In contrast to [Mean], where segmentation consistency is forced by a constraint, [Fixed] explicitly optimizes consistency in the loss function with a metric of spatial correspondence between the target segmentation label and the warped moving segmentation prediction $\mathcal{L}_{cons}^F$. An overview of [Fixed] framework is provided in supplementary Figure~\ref{fig3}. The loss function for [Fixed] is fomulated as:
\begin{equation}\label{eq9}
\mathcal{L}^F = \mathcal{L}_{seg}^F +\alpha \mathcal{L}_{reg}^F +\beta \mathcal{L}_{def}^F  + \gamma \mathcal{L}_{cons}^F.
\end{equation}
To evaluate the sole value of the mean-space strategy, we used the same network configurations and loss metrics for [Mean] and [Fixed] framework, and adapted the [Fixed] for SVF-based deformation instead of the displacement-based deformation used in the literature \cite{li2019hybrid}. 

To study the variability of evaluation metrics due to small processing errors, two additional reference methods are compared to apply morphology operations to the binary segmentations predicted by [Mean]: (1) the [Dilation] method dilates the segmentations by one voxel, (2) the [Translation] method translates the segmentations by one voxel along the axis perpendicular to the axial plane.

\subsubsection{Evaluations}\label{sss:3.2.2}
We applied the [Mean] framework and the reference methods to analyze the \textit{paired-scans} and \textit{grouped-scans}. Paired samples t-tests were used to statistically assess the difference of performance between approaches. First, using the \textit{paired} test set, \textbf{segmentation accuracy} of the methods was compared as quantified by the Dice coefficient with the segmentation labels. 

Second, using the reproducibility dataset, we evaluated \textbf{processing bias} of the methods in producing semantic segmentations and imaging biomarkers when time-points were passed in reversed order. Specifically, we computed the Cohen's kappa coefficient ($\kappa$) \cite{landis1977measurement} as voxel-wise agreement between segmentations, and measured the percent of variation ($\epsilon \%$) between estimation and reversed estimation of tract-specific measures (FA and volume). For volume ($V$) measurement, $\epsilon=[V^{''}- V^{'}]/[0.5\times(V^{''}+ V^{'})]\times 100 \%$,
where $V^{'}$ indicates the measured volume at a certain time-point, and $V^{''}$ the measured volume with reverse inputs. Therefore, for [Fixed], $V^{'}$ was obtained by direct segmentation and $V^{''}$ was obtained by propagation of the segmentation from the other time-point. This therefore indicates the bias in longitudinal processing when for instance the analysis of follow-up images is informed by the baseline \cite{reuter2012within}. For [Dilation], $V^{'}$ was the result of [Mean] and $V^{''}$ was the result of [Dilation]. 

Third, we evaluated the \textbf{geometric centrality} of the proposed [Mean] framework by visualizing the implicit reference image and the velocity fields. The norm of the average velocity fields $||\bar{\boldsymbol{v}}||_2^2$ was measured over the \textit{grouped} test set. 

Fourth, using the same dataset we compared the \textbf{consistency of segmentation} between [Mean] and [Fixed] methods. In the proposed [Mean] framework, input images were automatically aligned to the mean-space for joint segmentation, in which segmentations can be directly compared and designed to be consistent. For the [Fixed] framework, we utilized the same transformations by [Mean] to align its results to the mean-space for overlap evaluation (Dice). To reduce detrimental effects of resampling the [Fixed] results, we performed linear interpolation on probabilistic results, then binarized them and computed the Dice coefficient.

\subsubsection{Implementation}\label{sss:3.2.3}
Experiments were performed on an NVIDA 1080Ti GPU and an AMD 1920X CPU. Models were implemented using Keras-2.2.0 with a Tensorflow-1.4.0 backend, and trained using the Adam optimizer with a learning rate of $1 e^{-4}$. The order of input images was previously randomized for this study. In the experiments with paired inputs, models were trained bidirectionally (i.e., $5175\times2=10350$ and $3720\times2=7440$ transformations per epoch). For grouped inputs, we randomly shuffled image order in each training batch (size $=2$) for the group-wise [Mean] method. The pair-wise [Fixed] method was trained on all possible scan pairs within the training dataset of \textit{grouped-scans}, i.e., $1240\times3=3720$ pairs.

The hyperparameters in the loss function were empirically tuned based on the performance on the validation dataset. Specifically, for [Mean] algorithm, we used $\lambda_1=0.01$, and increased the $\lambda_2$ from $0.1$ to $0.5$ by $0.01$ per epoch. For the [Fixed] method, hyperparameters were set to $\alpha=10$, $\beta=0.1$ and $\gamma=1$. The training procedure was stopped until the validation loss tended to increase. The parameters with the best validation performance was selected for evaluation of the algorithms. 

\subsection{Results} \label{ss:3.4} 
The proposed mean-space joint framework ([Mean]) led to a significantly better spatial correspondence of segmentation ($\kappa=89.30\%$) than the fixed-reference framework ([Fixed], $\kappa=83.21\%$)(Table~\ref{tab1}). Typically, a $\kappa > 60 \%$ indicates “substantial” agreement, and a $\kappa > 80 \%$ indicates “almost perfect” agreement \cite{landis1977measurement}. This suggests that the spatial segmentation obtained by proposed framework is highly reliable regardless of order of input images, and the remaining small bias is even smaller than translation of one voxel which can easily happen due to subtle numerical instabilities and interpolation. This reliability is also corroborated by the subtle variation in determined imaging biomarkers. The percent variation of tract-specific volume, FA and MD measures ($0.13\%$, $0.04\%$ and $-0.01\%$) all had non-significant difference from zero. The segmentation accuracy of the proposed [Mean] framework did not improve over the [Fixed] method. Since the [Mean] framework aims at reducing variability in cross-sectional segmentations, it was indeed expected that overlap with the segmentation label \cite{de2015tract} at each time-point can decrease.

\begin{figure}[!ht]
\begin{center}
\includegraphics[width=0.9\textwidth]{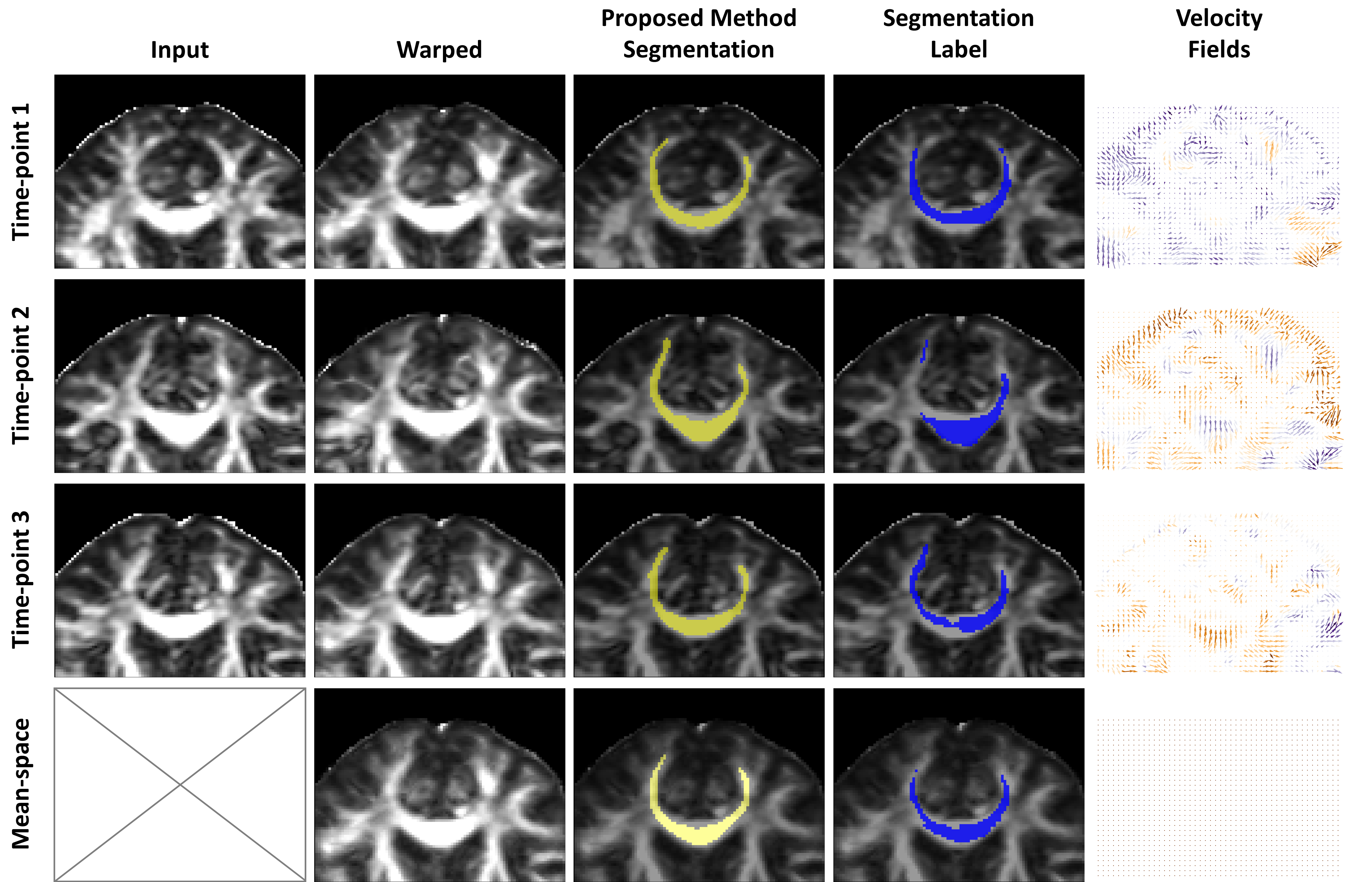}
\end{center}
\caption{\label{fig2}
Example registration and segmentation results (yellow) of the proposed [Mean] method in the native- and implicit mean-space of the inputs. For velocity fields the images are downsampled by 2 and the magnitude is scaled by $2.5$ for visualization only; the color denotes the direction of the perpendicular axis.}
\end{figure}

The bottom row of Fig.\ref{fig2} shows that adding three velocity fields gave a total deformation of zero at each corresponding location ( $||\bar{\boldsymbol{v}}||_2^2=2.98 \times e^{-16}$), indicating that the reference space was indeed exactly at the center of the deformations. Visual inspection of the warped FA images and their median image in the mean-space showed plausible average anatomy among the inputs. The segmented FMI tract appeared correct at all time-points, e.g., the under-segmentation of time-point 1 can be reasonable considering the follow-up images. Comparing with the consistent segmentation achieved by the proposed [Mean] framework using group-wise registration (Dice $=100\%$), the segmentation consistency of the pair-wise [Fixed] approach was suboptimal. The average Dice of aligned segmentations was $69.71 \pm 10.52 \%$ over the \textit{grouped} test set.
\begin{table}[!ht]
\caption{Processing bias and segmentation accuracy (Dice) of the proposed [Mean] framework compared with the [Fixed], [Dilation] and [Translation] methods. $\kappa$, Cohen's kappa. $\epsilon \%$, variation in measurements. \textbf{Bold font} indicates a significant improvement ($p<0.01$). \textbf{*} indicates non-significant difference from zero ($p>0.05$).} \label{tab1}
\begin{center}
\begin{tabular}{c|c|c|c|c|c}
\hline
 & \multicolumn{4}{|c|}{Processing bias}  & Accuracy \\
\hline
Method & $\epsilon (\%)$ in volume & $\epsilon (\%)$ in FA &$\epsilon (\%)$ in MD & $\kappa (\%) $ & Dice (\%)\\
\hline
[Dilation]  & $53.01 \pm 2.84$ & $-8.63 \pm 1.08 $  & $1.41 \pm 0.69 $ & $73.15 \pm 1.41 $ &$55.35 \pm 5.68 $\\

[Translation]  & - & $0.45 \pm 0.53$ & $-0.04^{\textbf{*}} \pm 0.63 $ & $84.77 \pm 1.49 $ &$61.68 \pm 7.53 $\\
\hline
\hline
[Fixed] & $0.70^{\textbf{*}} \pm 4.03 $ &$- 0.07^{\textbf{*}} \pm 1.03$ & $-0.04^{\textbf{*}}\pm 0.67 $ & $83.21 \pm 3.54$ & $\textbf{65.16} \pm \textbf{7.57}$\\

[Mean]  & $0.13^{\textbf{*}} \pm 2.33 $ &$0.04^{\textbf{*}} \pm 0.67$ & $- 0.01^{\textbf{*}} \pm 0.45$& $\textbf{89.30} \pm \textbf{2.34}$ & $64.46 \pm 7.51 $\\
\hline
\end{tabular}%
\end{center}
\end{table}

\section{Discussion and conclusion}
\label{s:4}

We presented a learning-based analytical framework for unbiased group-wise registration and consistent segmentation. In evaluations on a large-scale dataset, we showed that the proposed method was able to simultaneously align images in the mean-space of the group, and to produce highly reliable spatial segmentations and imaging biomarkers. It therefore can be used as a fully automated and integrated pipeline to support longitudinal imaging studies on large datasets. To our knowledge, this work advances the stage-of-the-art by the first time using constraint optimization for learning mean-space registration. We expect the proposed mean-space learning strategy can be widely applied to registration tasks.

This work focuses on longitudinal imaging studies of normal individuals, in which the number of acquired time-points is within a small range. For this, we assume that training models for a specific number of visits is acceptable. While for registration of a large number of images, the method is not yet applicable. Our experiments highlight the need for further research in developing evaluation criteria for segmentation accuracy and reliability. We used the standard Dice measure for each scan, but this measurement does not capture all the nuances of the resulting segmentation such as the consistency within-subject. 

For future extension of the method, we plan to analyze the influence of the weight between loss terms and to apply the method to analyze brain diseases with large and progressive changes, which may benefit from the proposed joint framework.






\acknowledgments 
This work was sponsored through grants of the Medical Delta Diagnostics 3.0: Dementia and Stroke, the Netherlands CardioVascular Research Initiative (Heart-Brain Connection: CVON2012-06, CVON2018-28), and the Dutch Heart Foundation (PPP Allowance, 2018B011).

\bibliography{spie.bib} 
\bibliographystyle{spiebib} 

\section*{SUPPLEMENTARY MATERIAL}

    \begin{figure} [ht]
    \begin{center}
    \includegraphics[width=\textwidth]{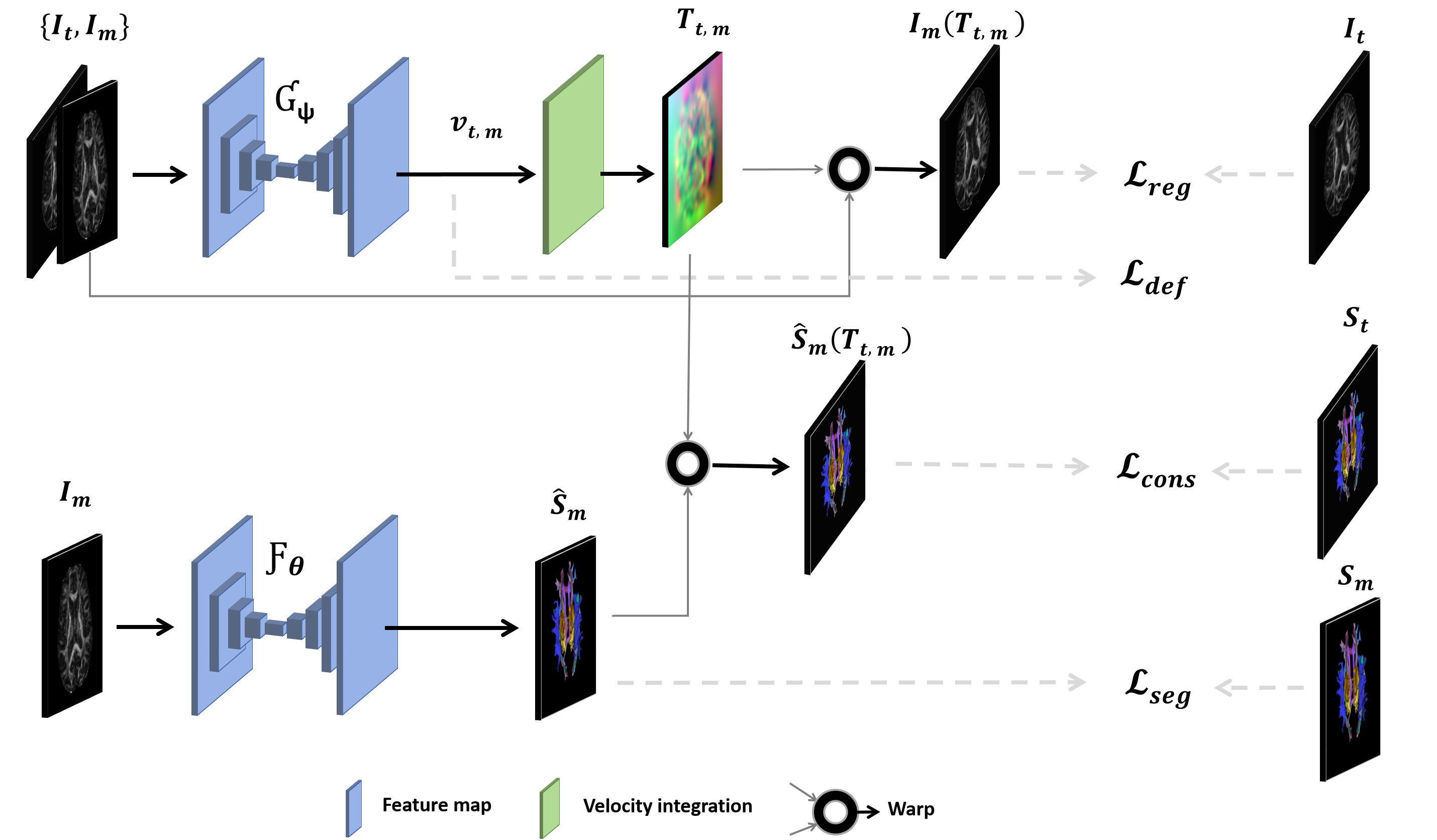}
    \setlength{\belowcaptionskip}{-15pt}
    \end{center}
\caption{\label{fig3}
Pipeline of the compared pair-wise joint optimization framework that uses fixed-reference registration ([Fixed]) \cite{li2019hybrid}. $I_t$ and $S_t$ indicate a target image and its segmentation image. Similarly, $I_m$ and $S_m$ are those for a moving image. $\mathcal{G}_{\boldsymbol{\Psi}}$ and $\mathcal{F}_{\boldsymbol{\Theta}}$ denote the registration and segmentation network, respectively. The loss function consists of $\mathcal{L}_{reg}$, $\mathcal{L}_{def}$, $\mathcal{L}_{cons}$ and $\mathcal{L}_{seg}$ terms.}
    \end{figure}

\end{document}